\documentclass[11pt]{article}
\usepackage{coling2020}
\usepackage{times}
\usepackage{url}
\usepackage{latexsym}

\usepackage{graphicx}
\usepackage{amsmath}

\usepackage{amssymb}
\usepackage{latexsym}
\usepackage{subfigure}
\usepackage{longtable}
\usepackage{booktabs}

\colingfinalcopy 

\title{CharBERT: Character-aware Pre-trained Language Model}

\author{
Wentao Ma$^\dag$,
Yiming Cui$^\ddag$$^\dag$,
Chenglei Si$^\P$$^\dag$,
Ting Liu$^\ddag$,
Shijin Wang$^\dag$$^\S$,
Guoping Hu$^\dag$\\
{$^\dag$State Key Laboratory of Cognitive Intelligence, iFLYTEK Research, China} \\
{$^\ddag$Research Center for Social Computing and Information Retrieval (SCIR), } \\
{Harbin Institute of Technology, Harbin, China} \\
{$^\S$iFLYTEK AI Research (Hebei), Langfang, China} \\
{$^\P$University of Maryland, College Park, MD, USA} \\
$^\dag$$^\S$\tt\{wtma,ymcui,clsi,sjwang3,gphu\}@iflytek.com \\
$^\ddag$\tt\{ymcui,tliu\}@ir.hit.edu.cn
}

\date{}

\begin{document}
\maketitle
\begin{abstract}
Most pre-trained language models (PLMs) construct word representations at subword level with Byte-Pair Encoding (BPE) or its variations, by which OOV (out-of-vocab) words are almost avoidable.
However, those methods split a word into subword units and make the representation incomplete and fragile.
In this paper, we propose a character-aware pre-trained language model named \textbf{CharBERT} improving on the previous methods (such as BERT, RoBERTa) to tackle these problems.
We first construct the contextual word embedding for each token from the sequential character representations, then fuse the representations of characters and the subword representations by a novel heterogeneous interaction module. 
We also propose a new pre-training task named NLM (Noisy LM) for unsupervised character representation learning.
We evaluate our method on question answering, sequence labeling, and text classification tasks, both on the original datasets and adversarial misspelling test sets.
The experimental results show that our method can significantly improve the performance and robustness of PLMs simultaneously.\footnote{Pretrained models, evaluation sets, and code are available at \url{https://github.com/wtma/CharBERT}}
\end{abstract}

\section{Introduction}
Unsupervised pre-trained language models like BERT \cite{devlin-2019-bert} and RoBERTa \cite{liu-2019-roberta} have achieved surprising results on multiple NLP benchmarks.
These models are pre-trained over large-scale open-domain corpora to obtain general language representations and then fine-tuned for specific downstream tasks.
To deal with the large vocabulary, these models use Byte-Pair Encoding (BPE) \cite{sennrich-2016-neural} or its variations as the encoding method.
Instead of whole words, BPE performs statistical analysis of the training corpus and split the words into subword units, a hybrid between character- and word-level representation.

\begin {figure} [t]
  \centering
  \includegraphics [width= 0.49\textwidth] {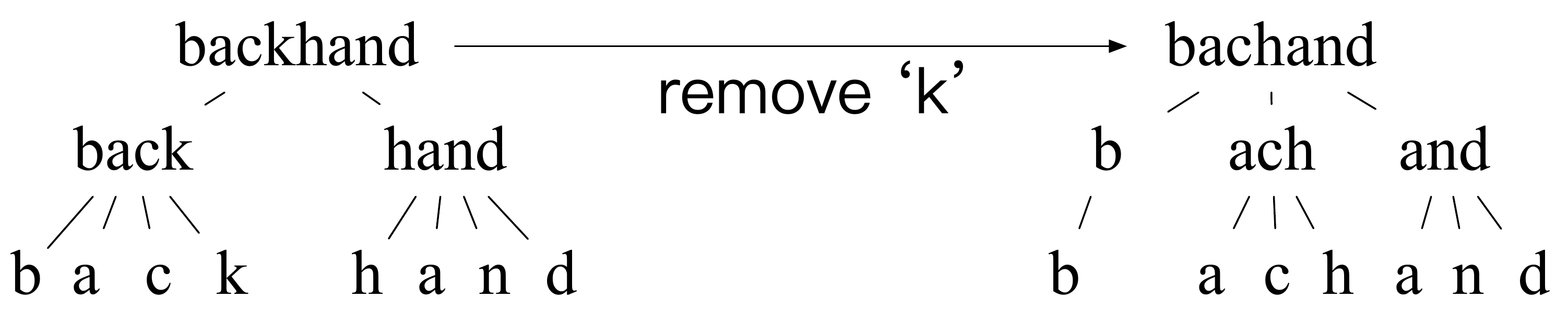}
  \caption{\label{charbert-example} The internal structure tree of \emph {backhand}, which has two subwords: \emph {back,hand}. If the letter  \emph {k} is removed, the subwords will change to \emph {b, ach, and}.}
\end {figure}

Even though BPE can encode almost all the words in the vocabulary into WordPiece tokens without OOV words, it has two problems:
1) incomplete modeling: the subword representations may not incorporate the fine-grained character information and the representation of the whole word;
2) fragile representation: minor typos can drastically change the BPE tokens, leading to inaccurate or incomplete representations. This lack of robustness severely hinders its applicability in real-world applications.
We illustrate the two problems by the example in Figure \ref {charbert-example}.
For a word like \emph {backhand}, we can decompose its representation at different levels by a tree with a depth of 3:
the complete word at the first layer, the subwords at the second layer, and the last characters.
BPE only considers representations of subwords on the second layer and misses the potentially useful information at the first and last layer.
Furthermore, if there is noise or typo in the characters (e.g., missing the letter `k'), the subwords and its number at the second layer will be changed at the same time. Models relying purely on these subword representations thus suffer from this lack of robustness. 

We take the CoNLL-2003 NER development set as an example. Nearly 28\% of the nouns words will be split into more than one subword with BERT tokenizer. When we randomly remove a character from the noun words in the dataset like the example in Figure \ref {charbert-example}, about 78\% of the words will be tokenized into completely different subwords, and 77\% of the words have a different number of subwords.

If we focus on the leaf nodes in the example, we can find that the difference of the two trees is only one leaf.  
So we extend the pre-trained language models by integrating character information of words. 
There are two challenges for character integration: 
1) how to model character information for whole words instead of subwords; 
2) how to fuse the character representations with the subwords information in the original pre-trained models.

We propose a new pre-training method CharBERT (BERT can also be replaced by other pre-trained models like RoBERTa) to solve these problems.
 Instead of the traditional CNN layer for modeling the character information, we use the context string embedding \cite{akbik-2018-contextual} to model the word's fine-grained representation.
 We use a dual-channel architecture for characters and original subwords and fuse them after each transformer block.
 Furthermore, we propose an unsupervised character learning task, which injects noises into characters and trains the model to denoise and restores the original word.
 The main advantages of our methods are: 
 1) character-aware: we construct word representations from characters based on the original subwords, which greatly complements the  subword-based modeling.
 2) robustness: we improve not only the performance but also the robustness of the pre-trained model;
 3) model-agnostic: our method is agnostic to the backbone PLM like BERT and RoBERTa, so that we can adapt it to any transformer-based PLM.
 In summary, our contributions in this paper are:
  \begin {itemize}
 \item We propose a character-aware pre-training method \textbf{CharBERT}, which can enrich the word representation in PLMs by incorporating features at different levels of a word; 
 \item We evaluate our method on 8 benchmarks, and the results show that our method can significantly improve the performance compared to the strong BERT and RoBERTa baselines;   
 \item We construct three character attack test sets on three types of tasks. The experimental results indicate that our method can improve the robustness by a large margin.
 \end {itemize}

\section{Related Work}

\noindent\textbf{Pre-trained Language Model. }
Early pre-trained language models  (PLMs) like CoVe \cite{mccann-2017-learned} and ELMo \cite {peters-elmo-2018} are pre-trained with RNN-based models, which are usually used as a part of the embedding layer in task-specific models.
GPT~\cite{GPT} used the transformer decoder for language modeling by generative pre-training and fine-tuned for various downstream tasks. 
BERT \cite{devlin-2019-bert} pre-trains the transformer encoder and uses self-supervised pre-training on the larger corpus, achieving surprising results in multiple natural language understanding (NLU) benchmarks.
Other PLMs such as RoBERTa \cite{liu-2019-roberta}, XLNet \cite{yang-2019-xlnet},  ALBERT \cite {lan-2019-albert} and ELECTRA \cite{clark-2019-electra}, improve on previous models  with various improvements on the model architectures, training methods or pre-training corpora.

To handle the large vocabularies in natural language corpora, most PLMs process the input sequence in subword units by BPE \cite{sennrich-2016-neural} instead of whole words, split a word into subwords by the byte pair encoding compression algorithm.
The size of BPE vocabulary usually ranges from 10K-100K subword units,  most of which are Unicode characters. 
Radford et al. \shortcite{radford-2019-language} introduce another implementation that uses bytes instead of Unicode characters as the base subword units, allowing BPE to encode any input sequence without OOV words with a modest vocabulary size (50K). 

\noindent\textbf{Character Representation. } 
Traditional language models employ a pre-defined vocabulary of words, but they cannot handle out-of-vocabulary words well.
Character language models (CLMs) can mitigate this problem by using a vocabulary of characters and modeling the character distribution for language modeling \cite{sutskever-2011-generating}.
CLMs have been shown to perform competitively on various NLP tasks, such as neural machine translation \cite{lee-2017-fully} and sequence labeling \cite{csahin-2018-character,akbik-2018-contextual}. 
Furthermore, character representation has also been used to construct word representation; for example, Peters et al. \shortcite {peters-elmo-2018} construct the contextual word representation with character embeddings and achieve significant improvement.

\noindent\textbf{Adversarial Attack. }
PLMs are fragile to adversarial attacks, where human-imperceptible perturbations added to the original examples fool models to make wrong predictions. 
Jia and Liang \shortcite {jia-2017-adversarial} and Si et al. \shortcite{Si2020BenchmarkingRO} show that state-of-the-art reading comprehension models can be fooled even with black-box attacks without accessing model parameters. 
Other white-box attacks \cite{geneticAttack,PWWS,TextFooler,SememePSO} use gradients or model prediction scores to find adversarial word substitutes as effective attacks.
For character-level attacks, Belinkov and Bisk \shortcite {belinkov-2017-synthetic} studied how synthetic noise and noise from natural sources affect character-level machine translations.
Ebrahimi et al. \shortcite {ebrahimi-2018-adversarial}  investigated adversarial examples for character-level neural machine translation with a white-box adversary.
To defend against character-level attacks, Pruthi et al. \shortcite {pruthi-2019-combating} propose to place a word recognition model before the downstream classifier to perform word spelling correction to combat spelling mistakes.

\noindent\textbf{Heterogeneous Representation Fusion.} In our work, we need to fuse heterogeneous representations from two different sources. Similar modules have been applied before under different settings such as machine reading comprehension \cite{BIDAF,gating} and pre-trained language models~\cite{ernie,SemBERT}. Different from these works, we design a two-step fusion module to fuse the character and subword representations by a interactive way, which can be extended to integrate other information into language model (e.g. diacritics or external knowledge). 

\section{Methodology} 
In this section, we present the overall framework of CharBERT and its submodules, including the model architecture in Section 3.2, the character encoder in Section 3.3, the heterogeneous interaction module in Section 3.4, the new pre-training task in Section 3.5, and the fine-tuning method in Section 3.6.

\subsection{Notations} 
We denote an input sequence as $\{w_1, ..., w_i, ..., w_m\}$, where $w_i$ is a subword tokenized by BPE and $m$ is the length of the sequence in subword-level.
Each token $w_i$ is consisted of characters $\{c_1^i,..., c_{n_i}^i\}$ and $n_i$ is the subword's length. 
We denote the length of input in character-level as $N$, where $N=\sum_{i=1}^mn_i$.

\subsection{Model Architecture} 
\begin {figure*} [t]
  \centering
  \includegraphics [width= 0.8\textwidth] {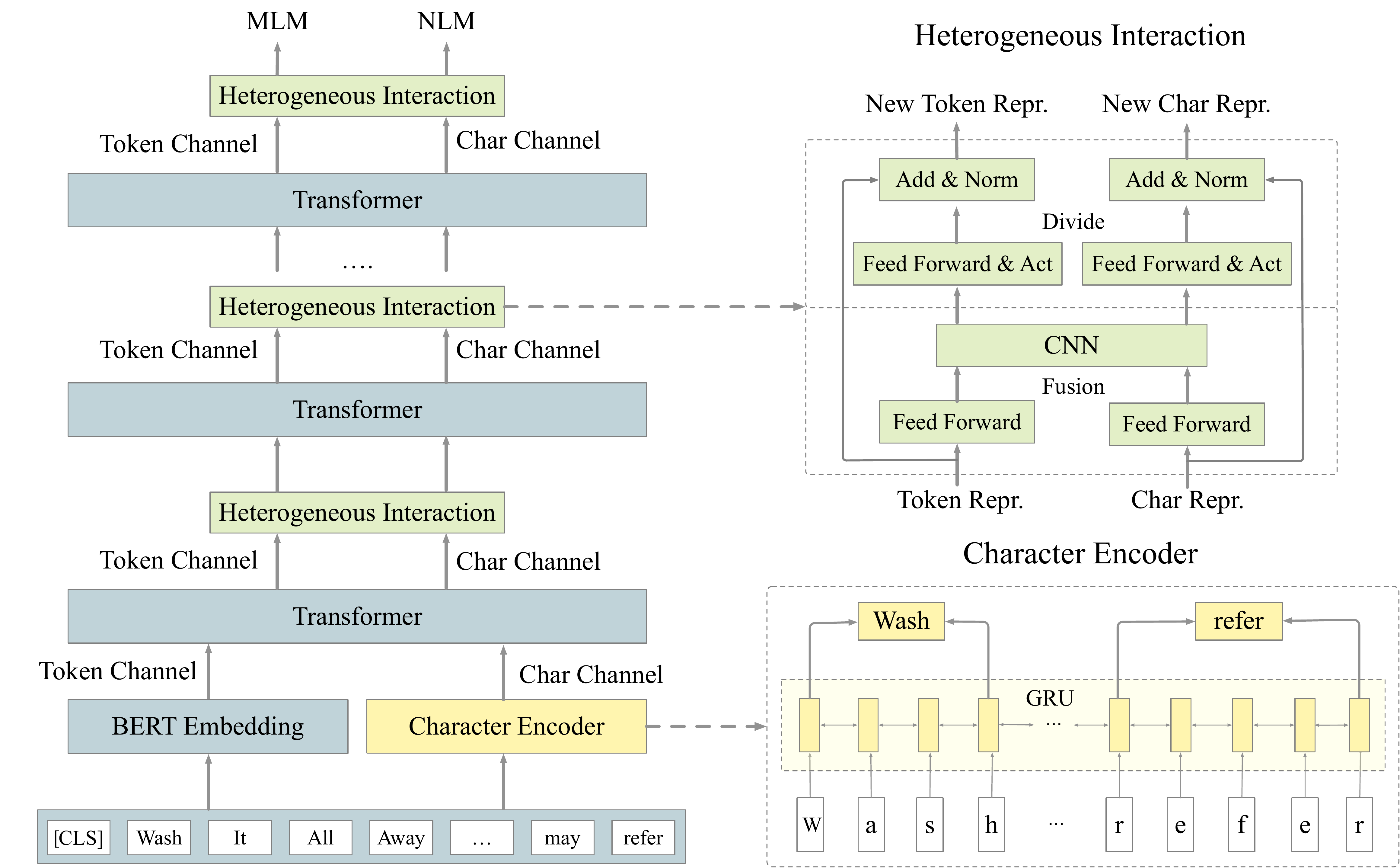}
  \caption{\label{charbert-arch} The neural architecture of CharBERT. The left part is the main structure of CharBERT based on the original pre-trained models like BERT. The modules in the right part are the heart of CharBERT: character encoder and heterogeneous interaction. \emph{(best viewed in color)}}
\end {figure*}

As shown in Figure \ref{charbert-arch}, we use a dual-channel architecture to model the information from subwords and characters, respectively.
Besides the transformer layers from the original pre-trained model like BERT, the core modules of CharBERT are: 
1) the character encoder, responsible for encoding the character sequence from the input tokens;
2) heterogeneous interaction,  fuse the information from the two sources and construct new independent representations for them.

We model the input words as sequences of characters to catch the character information within and among subwords, a supplement for WordPiece embedding.
The character-level representation is heterogeneous with subword-level representation from the embedding layer of pre-trained models as they come from different sources. However, they capture information at the different granularity and complement each other. 
In order to enable them to enrich each other effectively, we design a heterogeneous interaction module with two steps:  
1) fuse: fuse the information from the dual-channel based on the CNN layer \cite{kim-2014};
2) split: build new representations for each channel based on residual connection.

\subsection{Character Encoder} 
In this module, we need to form token-level embeddings with the input sentences as sequences of characters.
We first convert the sequences of tokens into characters and embed them into fixed-size vectors.
We then apply a bidirectional GRU layer \cite{cho-2014-learning} to construct the contextual character embeddings, which can be formulated by
\begin {gather}
\textstyle e_j^i = W_c  \cdot c_j^i  ~~ ;~~ \textstyle {h_j^i(x)} = {\text{Bi-GRU}}(e_j^i);
\end {gather}
where $W_c$ is the character embedding matrix , $h_j^i$ is the representation for $j$th character in the $i$th token. 
We apply the bi-GRU on the characters with a length of $N$ for the whole input sequence instead of a single token, building the representations from the characters within and among the subwords.
To construct token-level embeddings, we concatenate the hidden of the first and last character of the token.
\begin {gather}
\textstyle h_i(x) = [{h_1^i(x)}; {h_{n_i}^i(x)} ]
\end {gather}
where $n_i$ is the length of $i$th token and $h_i(x)$ is the token-level embedding from characters.
The contextual character embeddings are derived by characters and can also catch the full word information by bi-GRU layers.

\subsection{Heterogeneous Interaction} 
The embeddings from characters and the original token-channel are fed into the same transformer layers in pre-trained models.
The token and char representations are fused and split by the heterogeneous interaction module after each transformer layer.

In the fusion step, the two representations are transformed by different fully-connected layers.
Then they are concatenated and fused by a CNN layer, which can be formulated by
\begin {gather}
\textstyle t'_i(x)= W_1 * t_i(x) + b_1 ~~;~~ \textstyle h'_i(x)= W_2 * h_i(x) + b_2 \\
\textstyle w_i(x) = [t'_i(x); h'_i(x)] ~~;~~ \textstyle m_{j,t} = tanh(W_3^j * w_{t:t+s_j-1} + b_3^j)
\end {gather}
where $t_i(x)$ is the token representations, $W$, $b$ are parameters, $w_{t:t+s_j-1}$ refers to the concatenation of the embedding of ($w_t$,...,$w_{t+s_j-1}$), $s_j$ is the window size of $j$th filter, and $m$ is the fusion representation with the dimension same with the number of filters.

In the divide step, we transform the fusion representations by another fully connected layer with GELU activation layer \cite {hendrycks-2016-gaussian}. 
We then use the residual connection to retain the respective information from the two channels.
\begin {gather}
\textstyle m_i^t(x) = \delta( W_4 * m_i(x) + b_4) ~~;~~ \textstyle m_i^h(x) =\delta(  W_5 * m_i(x) + b_5) \\
\textstyle T_i(x) =  \ t_i(x) + m_i^t(x)~~ ;~~ \textstyle H_i(x) = h_i(x) + m_i^h(x)
\end {gather}
Where $\delta$ is the activation function GELU, $T$ and $H$ is the new representations of the two channels.
To prevent vanishing or exploding of gradients, a layer normalization \cite{ba-2016-layer} operation is applied after the residual connection.
 
 By the fusion step, the representations from the two channels can enrich each other.
 By the divide step, they can keep their unique features from token and character, and learn the different representations in dual-channel by their own pre-training tasks.
 
 \begin {figure*} [t]
  \centering
  \includegraphics [width= 0.8\textwidth] {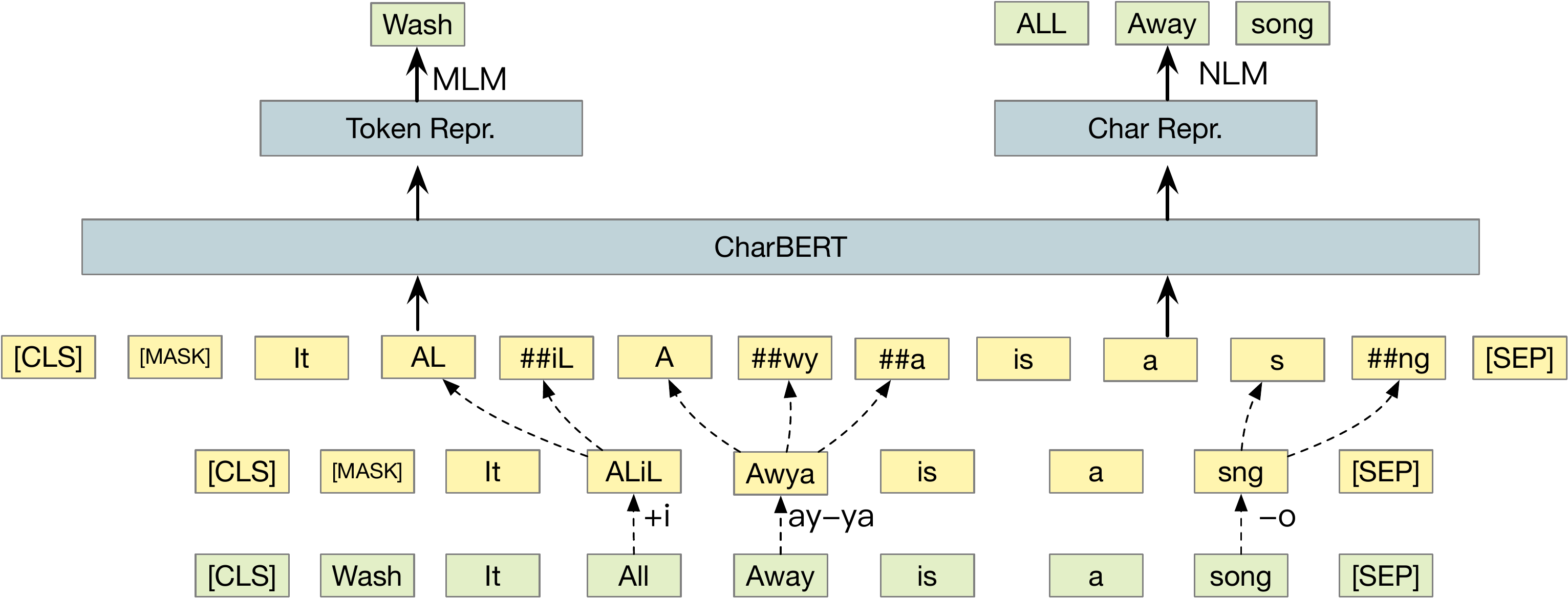}
  \caption{\label{charbert-nlm} Character-aware language model pretraining. The MLM task is similar to the one in BERT, but with lower mask probability (10\%). The NLM task introduces the character noises by dropping,  adding and swapping internal characters within the word and predicts the original whole word by the representation from the character channel. }
\end {figure*}
 
\subsection{Unsupervised Character Pre-training} 
To learn the representation from the internal morphological feature within the words, we propose an unsupervised character pre-training task named noisy language modeling (NLM) for CharBERT.
We introduce some character noises into the words, and predict the original words by the representations from the character channel as shown in Figure \ref {charbert-nlm}.

Following the previous work \cite {pruthi-2019-combating}, we change the original character sequence by dropping, adding, and swapping internal characters within the whole word.
As the number of subwords may be changed after introducing the noise, the objective of the pre-training tasks is to predict the whole original word instead of subwords.
We construct a new word-level vocabulary as the prediction space
\begin {gather}
\textstyle H'_i = \delta( W_6 * H_i + b_5) ~~ ;~~ \textstyle p(W_j | H_i) = \frac {\exp(linear(H'_i) \cdot W_j) }{ \sum_{k=1}^S \exp(linear(H'_i) \cdot W_k)}
\end {gather}
where $linear(\cdot)$ is a linear layer, $H_i$ is the token representations from the character channel, $S$ is the size of the word-level vocabulary.

Similar to BERT, CharBERT also adopts masked language modeling (MLM) as the pre-training task for the token channel.
Different from NLM , MLM enables CharBERT to capture lexical and syntactic information in token-level.
Note that, we only mask or replace the tokens without any character noise for MLM. 
More details of the pre-training tasks can be found in \newcite{devlin-2019-bert}.

\subsection{Fine-tuning for Specific Tasks} 
Most of the natural language understanding tasks can be simply divided into two groups: token-level tasks like sequence labeling and sequence-level tasks, such as the text classification tasks.
For token-level tasks, we concatenate the final output embeddings from the two channels in CharBERT as the input for fine-tuning.
For sequence-level tasks, most of the pre-trained models use the representation of a special token like \texttt {[CLS]}   for prediction.
In this paper, to adequately take advantage of the character- and token-level information in the sequence, we perform average over all the embeddings after concatenating the representations from the two channels in the last layer of CharBERT for sequence level classification.

\section{Experiments} 
In this section, we present the pre-training details of CharBERT and the fine-tuning results on three kinds of tasks: question answering, sequence labeling, and text classification.
Furthermore, we construct three character attack test set from those tasks and evaluate the robustness of CharBERT.

\subsection{Experimental Setup}
We use BERT \cite{devlin-2019-bert} and RoBERTa  \cite{liu-2019-roberta} base as our main baseline, where the models consist of 12 transformer layers, with 768 hidden size and 12 attention heads.
The vocabulary of BERT and RoBERTa contains 30K and 50K subword units respectively,  and the total parameters of them are 110M and 125M.
The size of additional parameters for BERT and RoBERTa is 5M, which means the character channel is much smaller than the token channel in original pre-trained models.
We change 15\% of the input words for NLM and lower the mask probability from 15\% to 10\% in MLM task to avoid too much information loss in the sequence.

We use English Wikipedia (12G, 2,500M words) as our pre-training corpus and adopt the parameters of the pre-trained models to initialize the token channel of CharBERT.
In the pre-training step, we set the learning rate as 5e-5, batch size as 32, and pre-train CharBERT 320K steps.
The word-level vocabulary contains 30K words for NLM, and the size of the character vocabulary is 1000.
We use 2 NVIDIA Tesla V100 GPUs, with 32GB memory and FP16 for pre-training, which is estimated to take 5 days.
For fine-tuning, we find the following ranges of possible values work well on the downstream tasks, i.e., batch size 16, learning rate:  3e-5,  2e-5, number of epochs ranging from 2 to 6.

For the optimizer, we use the same setting with the pre-trained model in the token channel like BERT and RoBERTa, both in pre-training and fine-tuning steps.
For experimental comparison, we mainly compare CharBERT with previous state-of-the-art pre-trained models in BERT$_{\texttt{base}}$ setting. 
We will also pre-train CharBERT with pre-trained models in BERT$_{\texttt{large}}$ setting in the future.

 \begin{table}[t]
        \begin{center}\small
        \begin{tabular}{l cc | cc | cccc}
        \toprule
         &  \multicolumn{4}{c}{\bf  SQuAD } &  \multicolumn{4}{c}{\bf {Text Classification} }  \\
         Models&  \multicolumn{2}{c}{\bf    1.1} & \multicolumn{2}{c}{\bf { 2.0} } & \bf  CoLA & \bf  MRPC &\bf  QQP & \bf  QNLI \\ 
         &  EM &  F1 &   EM &   F1  &  Corr &  Acc & Acc & Acc    \\ 
        \midrule
        BERT \cite{devlin-2019-bert}  & 80.5 & 88.5  & 73.7 & 76.3 & 57.4 & 86.7  & 90.6 & 90.7 \\
        CharBERT & \bf 82.9 & \bf 89.9 & \bf 75.7 & \bf 78.6 & \bf 59.1 & \bf 87.8 & \bf 91.0 & \bf  91.7 \\ 
        \midrule
        RoBERTa \cite{liu-2019-roberta} & \bf  84.6 & \bf  91.5 & 80.5 & 83.7 & \bf62.1 & 90.2 & 91.2 & 92.8 \\
        XLNet \cite {yang-2019-xlnet}  &- &- & 80.2 & -  &   60.2 & 88.2 &  91.4 & 91.7  \\
        CharBERT$_{\texttt{RoBERTa}}$  & 84.0 & 90.9 & \bf 81.1 & \bf 84.5 & 61.8  & \bf 90.4 & \bf 91.6 & \bf 93.4 \\
        \bottomrule
        \end{tabular}
        \end{center}
        \caption{\label{result-squad} Experimental results of our model and previous strong pre-trained models under BERT$_{\texttt{base}}$ setting on the dev set of Question Answering and Text Classification tasks. We report exact match (EM) and F1 scores for SQuAD,  Matthew's correlation for CoLA, and accuracy for other tasks.}
         \end{table}

\subsection{Results on Question Answering (SQuAD)}
The Stanford Question Answering Dataset (SQuAD) task requires to extract the answer span from a provided passage based on specified questions.
We evaluate on two versions of the dataset: SQuAD 1.1 \cite {rajpurkar-2016-squad} and SQuAD 2.0 \cite {rajpurkar-2018-know}.
For any question in SQuAD 1.1, there is always one or more answers in the corresponding passage.
While for some questions in SQuAD 2.0, there is no answer in the passage.
In the fine-tuning step for SQuAD, we concatenate the outputs from the character and token channel from CharBERT and use a classification layer to predict whether the token is a start or end position of the answer.
For SQuAD 2.0, we use the probability on the token  \texttt {[CLS]} as the results of no answer and search the best threshold for it.

The results are reported on Table \ref {result-squad}. 
For comparable experiments, all of the results are reported by a single model without other tricks like data augmentation.
We can find that our character-aware models (CharBERT, CharBERT$_{\texttt{RoBERTa}}$) outperform the baseline pre-trained models except for RoBERTa in SQuAD 1.1, which indicates the character information probably can not help the remaining questions.
         
 \subsection{Results on Text Classification}
We select four text classification tasks for evaluation:  CoLA \cite{warstadt-2019-neural}, MRPC \cite{dolan-2005-automatically}, QQP, and QNLI \cite{wang-2018-glue}.
CoLA is a single-sentence task annotated with whether it is a grammatical English sentence.
MRPC is a similarity task consisted of sentence pairs automatically extracted from online news sources, with human annotations for whether the sentences in pairs are semantically equivalent.
QQP is a paraphrase task with a collection of question pairs from the community question-answering website Quora, annotated with whether a pair of questions are semantically equivalent.
QNLI is an inference task consisted of question-paragraph pairs, with human annotations for whether the paragraph sentence contains the answer.

The results are reported in Table \ref {result-squad}.
For the BERT based experiments, CharBERT significantly outperforms BERT in the four tasks.
In the RoBERTa based part, the improvement becomes much weaker for the stronger baseline.
We find that the improvement in text classification is weaker than the other two kinds of tasks, which may be because the character information contributes more to token-level classification tasks like SQuAD and sequence labeling.

\subsection{Results on Sequence Labeling}
\label{sec:results-ner}
To evaluate performance on token tagging tasks, we fine-tune CharBERT on CoNLL-2003 Named Entity Recognition (NER) \cite{sang-2003-introduction} and Penn Treebank POS tagging datasets.\footnote {\url{https://catalog.ldc.upenn.edu/LDC2015T13}}
CoNLL-2003 NER dataset consists of 300k words, which have been annotated as \emph {Person, Organization, Miscellaneous, Location,}  or \emph {Other}.
The POS tagging dataset comes from the Wall Street Journal (WSJ) portion of the Penn Treebank, containing 45 different POS tags and more than 1 million words.
For fine-tuning, we feed the representations from the dual-channel of CharBERT into a classification layer over the label set.
Following the setting in BERT \cite{devlin-2019-bert}, we use the hidden state corresponding to the first sub-token as input to the classifier.

The results in reported in Table \ref {result-seqlabel}. We introduce two strong baselines Meta-BiLSTM \cite{bohnet-2018-morphosyntactic} and Flair Embeddings \cite{akbik-2018-contextual} in the two tasks for comparison. 
Our model (CharBERT, CharBERT$_{\texttt{RoBERTa}}$) exceeds the baseline pre-trained models BERT and RoBERTa significantly (p-value $\leq 0.05$), and we set new state-of-the-art results on the POS tagging dataset.
 \begin{table}[t]
        \begin{center}\small
        \begin{tabular}{l cc | cc | cc}
        \toprule
         &  \multicolumn{2}{c}{\bf QNLI } & \multicolumn{2}{c}{\bf {CoNLL-2003 NER}  }  &  \multicolumn{2}{c}{\bf SQuAD 2.0 } \\
         Models &  Original   & Attack    & Original & Attack &  Original & Attack \\ 
        \midrule
        BERT & 90.7 & 63.4  &  91.24  & 60.79  & 76.3 & 50.1  \\
        AdvBERT & 90.8  & 75.8  & 90.68 & 71.47  & 76.6  & 52.4  \\
        BERT+WordRec & 84.0 & 76.1 & 82.52 & 67.79  & 63.5 & 55.2  \\
        CharBERT & \bf 91.7 &  \bf 80.1  & \bf 91.81 & \bf 76.14 & \bf 78.6  &   \bf 56.3   \\ 
        \bottomrule
        \end{tabular}
        \end{center}
        \caption{\label{result-advchar} Experimental results of robustness evaluation. We report accuracy for QNLI,  F1-score for CoNLL-2003 NER and SQuAD 2.0. We construct the `Attack' sets with `Original' ones by introducing four kinds of character-level noise.}
         \end{table}

\subsection{Robustness Evaluation}
\label {sec: robustness}
We conduct the robustness evaluation on adversarial misspellings with BERT based models.
Followed the previous work \cite {pruthi-2019-combating}, we use four kinds of character-level attack: 
1) dropping: drop a random character within the word;
2) adding: add a random character into the word;
3) swapping: swap two adjacent characters within the word;   
4) keyboard: replace a random internal char with a nearby char on the keyboard.
We only apply the attack perturbation on words with length no less than 4 and we randomly select one of the four attacks to apply on each word.

For the evaluation tasks, we consider all the three types of tasks: questioning answering, sequence labeling, and text classification.
That is different from the previous works on adversarial attack and defense \cite {ebrahimi-2018-adversarial,pruthi-2019-combating}, which usually focus only on a specific task like machine translation or text classification.
We select the SQuAD 2.0, CoNLL-2003 NER, and QNLI datasets for the evaluation.

For the dev set in SQuAD 2.0, we only attack the words in questions.
For CoNLL-2003 NER and QNLI, we attack all the words under the length constraint.
In this set-up, we modify 51.86\% of the words in QNLI, 49.38 \% in CoNLL-2003 NER, and 22.97\% words in SQuAD 2.0.
We compare our CharBERT model with three baselines: 1) the original BERT model; 2) BERT model with adversarial training (AdvBERT), which is pre-trained by the same data and hyper-parameters with CharBERT; 3) BERT with word recognition and pass-through back-off (BERT+WordRec), we use the pre-trained scRNN typo-corrector from \cite {pruthi-2019-combating}. All the inputs are `corrected' by the typo-corrector and fed into a downstream model. We replace any OOV word predicted by the typo-corrector with the original word for better performance. 

The results are reported in Table \ref{result-advchar}.
The performance of BERT drops more than 30\% on the misspelling test sets,  which shows that BERT is brittle for the character misspellings attack.
AdvBERT and BERT+WordRec have moderate improvement on the misspellings attack sets, compared to the BERT baseline.
We find that the performance of BERT+WordRec has dropped significantly in the original set due to the error recall for normal words.
In comparison, CharBERT has the least performance drop than the other baselines under the character attacks, which denotes that our model is the most robust for the misspellings attack in multiple tasks, while still achieving improvement on the original test sets at the same time. Note that AdvBERT was pre-trained on the same data for the same number of training steps as our CharBERT model, except that AdvBERT does not have our proposed new methods. Thus, the comparison between AdvBERT and CharBERT can highlight the advantages of our method. 

\begin{table}[t]
\begin{minipage}{\textwidth}
\begin{minipage}[t]{0.45\textwidth}
\centering
\small
 \begin{tabular}{l cc}
        \toprule
         & \bf  NER & \bf  POS  \\ 
        Models  &   F1-score &   Accuracy  \\ 
        \midrule
        MetaBiLSTM  & - & 97.96 \\
        Flair Embeddings & \bf 93.09 & 97.85 \\
        BERT  & 91.24 & 97.93 \\
        CharBERT &  91.81 & \bf 98.05 \\
        \midrule
        RoBERTa & 92.22 & 97.98 \\
        CharRERT$_{\texttt{RoBERTa}}$ & \bf 92.49 & \bf 98.09 \\
        \bottomrule
        \end{tabular}
\makeatletter\def\@captype{table}\makeatother\caption{\label{result-seqlabel} Experimental results of our model and previous strong models on the test set of CoNLL-2003 NER and WSJ Postag datasets.} 
\end{minipage}
\hspace{0.3cm}
\begin{minipage}[t]{0.50\textwidth}
\centering
\small
 \begin{tabular}{l cccc}
        \toprule
         & \bf  SQuAD 2.0 & \bf  NER &\bf  QNLI & \bf  QNLI-Att \\ 
         Models &  F1 &  F1 & Acc & Acc  \\ 
        \midrule
        CharBERT & \bf 78.6 & \bf 91.81  &  \bf   91.7 & \bf 80.1 \\
         ~ w/o GRU & 77.7   & 91.45   & 90.8  &  76.9  \\
         ~ w/o HI     & 76.8  &  91.28  & 90.9  & 77.7   \\
         ~ w/o NLM & 78.3   &  91.69  &91.4  & 68.3   \\
        \midrule
        AdvBERT    & 77.4  & 91.03   & 90.7  &   75.8 \\
       BERT  & 76.3 & 91.24 & 90.7 & 63.4 \\
        \bottomrule
        \end{tabular}
\makeatletter\def\@captype{table}\makeatother\caption{\label{result-ablation} Experimental results of ablation study. AdvBERT can be considered as CharBERT without  the three modules but in the same pre-training setting. }
\end{minipage}
\end{minipage}
 \end{table}

\subsection{Ablation Study}
We consider the three modules in CharBERT: character encoder, heterogeneous interaction, and the pre-training task NLM in the ablation experiments.
For character encoder, we remove the GRU layer and use the character embeddings as the character representation (w/o GRU).
For the heterogeneous interaction module, we remove the whole module and the two channels have no interaction with each other in the model  (w/o HI).
For the pre-training tasks, we remove NLM and concatenate the representations from the two channels in CharBERT for MLM in the pre-training step (w/o NLM).
At last, we also compare with the two baseline models AdvBERT and BERT.  
We can consider AdvBERT as a fair baseline model with the same weight initialization, training data and training steps as CharBERT, without our three proposed modules.
We select four tasks: SQuAD 2.0, CoNLL-2003 NER, QNLI, and QNLI with character attack (QNLI-Att) from the four parts of the experiments above for evaluation.

We can see the ablation results from Table \ref {result-ablation}.
When we remove the GRU layer or heterogeneous interaction module, the performance drops significantly in all the tasks.
While we remove NLM in the pre-training step, the model has a similar performance in SQuAD 2.0, NER, and QNLI tasks, but has a much worse performance in QNLI-Att set, which denotes that the pre-training task significantly improves the robustness of CharBERT.
Furthermore, CharBERT (w/o NLM) still has a much better performance than BERT, which means CharBERT has better robustness even without the pre-training task.

\section{Analysis} 

In this section, we conduct some experiments on CoNLL-2003 NER task with the test set to further analyze the `incomplete modeling' and `fragile representation' problems. In the end, we compare the contextual word embeddings generated by BERT and CharBERT with a feature-based method.

\subsection{Word vs. Subword} 
To find out the effect of `incomplete modeling' problem on the word representation,  we divide all the words in the dataset into `Word' and `Subword' groups by whether the word will be split into multiple subwords. In fact, the `Subword' group only has 17.8\% of words but has 45.3\% of named entities. 

The results of BERT and CharBERT are in Figure \ref {subwords-results}. For the results of the same model in different groups, we find that the performance in `Subword' group are significantly lower than the ones in `Word' group, which indicates that the representations based on subwords may be insufficient for the words. For the results of different models in the same group, the improvement of CharBERT in `Subword' group is 0.68\%, which is much higher than that in `Word' group (0.29\%). That means the main improvement comes from the `Subword' part, and CharBERT can generate better representations for the words with multiple subwords. In other words,  CharBERT can alleviate the `incomplete modeling' problem by catching the information among different subwords with the GRU layer.

\subsection{Robustness Analysis} 

In this part, we further explore how the contextual word embeddings change over the character noise.
Specifically, we need to find out whether the representations from CharBERT are more or less sensitive to changes in character level.
We define a metric to measure the sensitivity of pre-trained language models over a specific dataset
\begin {gather}
\textstyle S = \frac{1}{m}\sum_{i=1}^m(-\frac{1}{2}cos(h(t_i), h_i(t'_i)) + 0.5)
\end {gather}
where $cos$ is cosine similarity, m is the number of words in dataset, $h$ is the last hidden in the model, $t_i$ is the $i$th word in the set and $t'_i$ is the same word with character noise.
In extreme cases, if a model is not sensitive at all to the character attack, the two vectors would be the same, yielding S=0.

\begin{figure}[t]
\begin{minipage}[t]{0.312\textwidth}
\centering
\includegraphics[width=\textwidth]{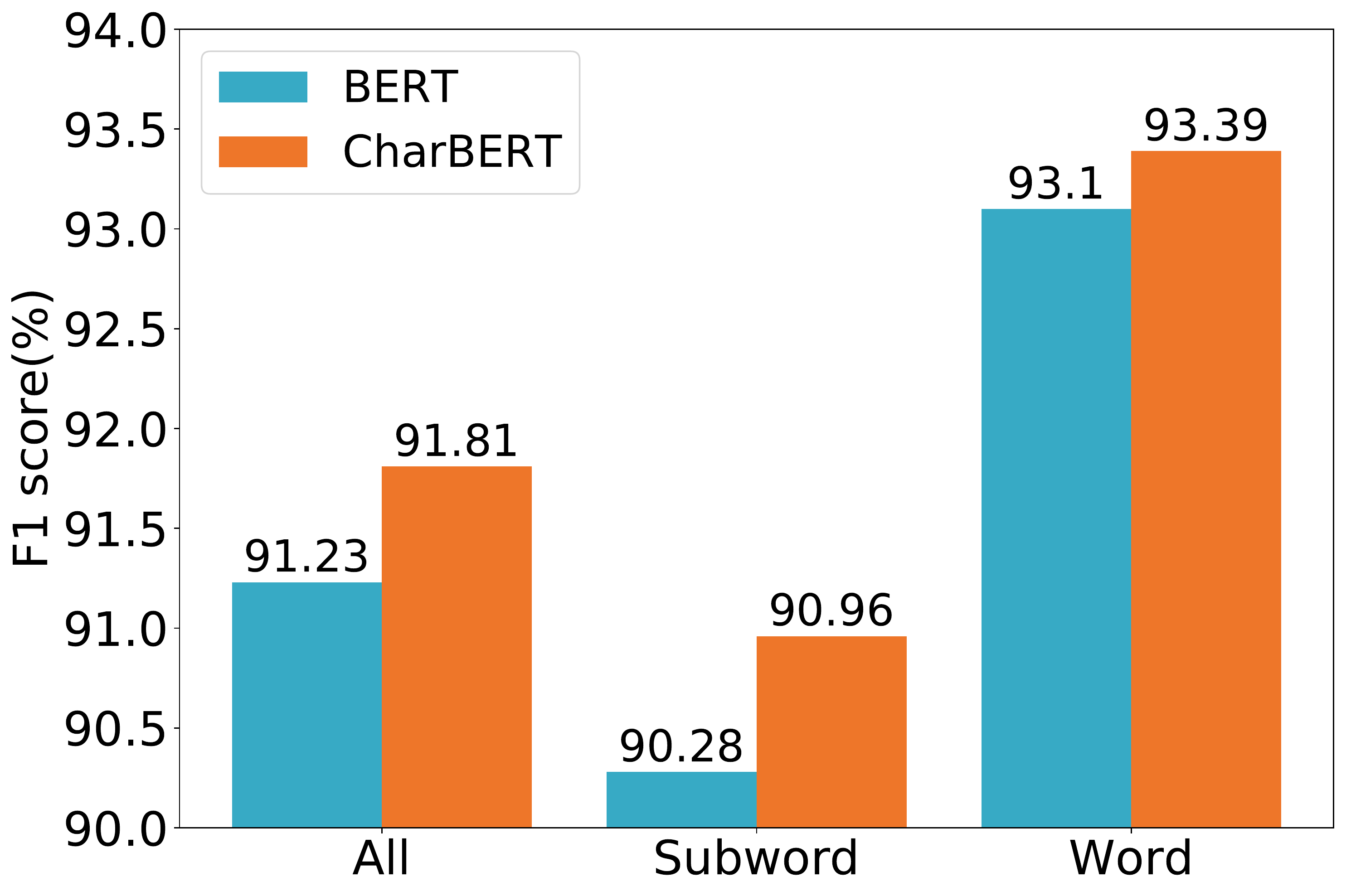} 
\caption{\label{subwords-results} The performances of different parts in ConNLL-2003 NER test set. `Subword' means the words will be split into more than one subwords.}
\end{minipage}
\hspace{0.1cm}
\begin{minipage}[t]{0.33\textwidth}
\centering
\includegraphics[width=\textwidth]{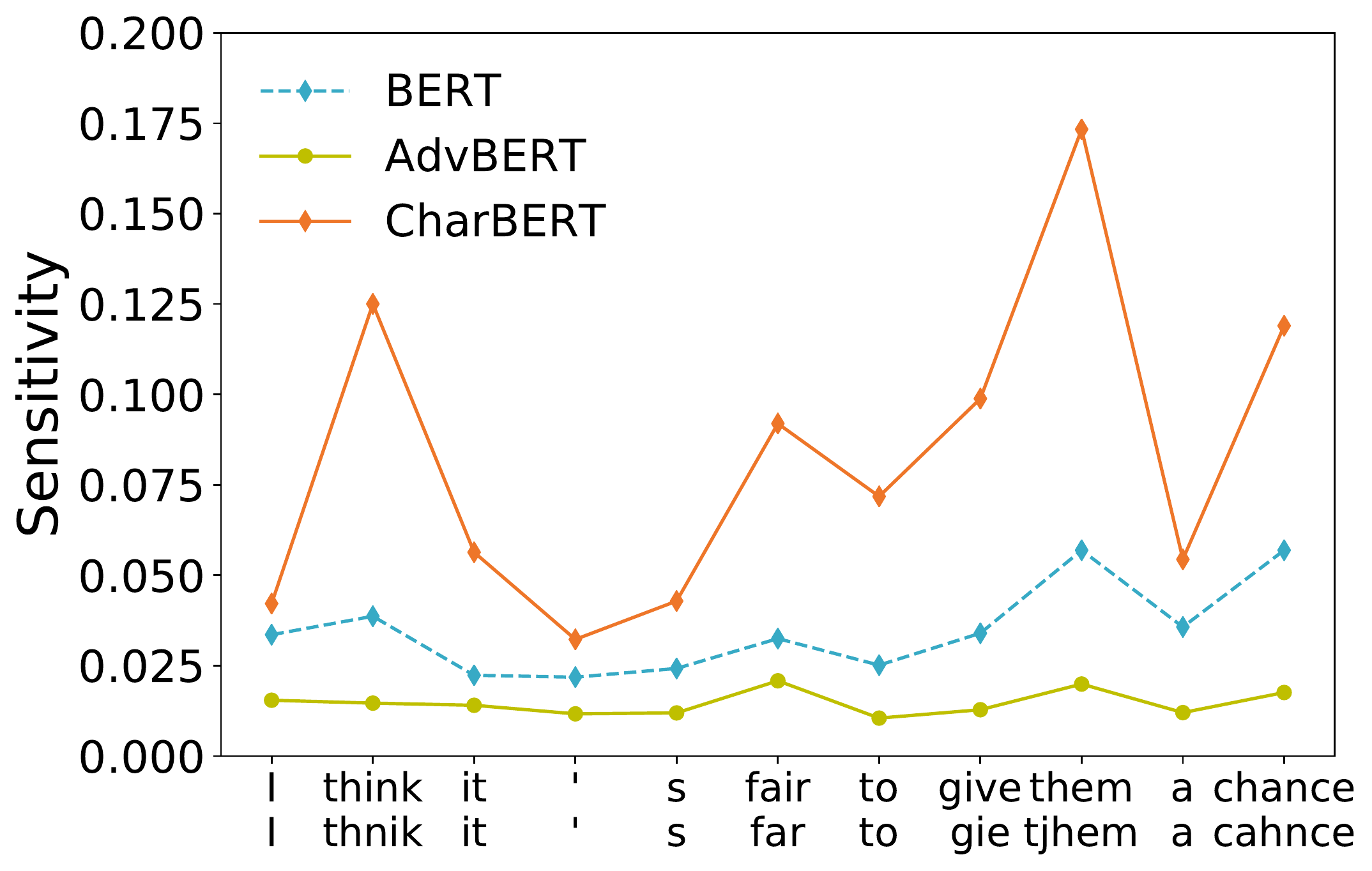}
\caption{\label{charbert-sen} The sensitivity result of a sample in CoNLL-2003 NER test set. The words `think, fair, give, them, chance' are changed in the attack set. }
\end{minipage}
\hspace{0.1cm}
\begin{minipage}[t]{0.325\textwidth}
\centering
\includegraphics[width=\textwidth]{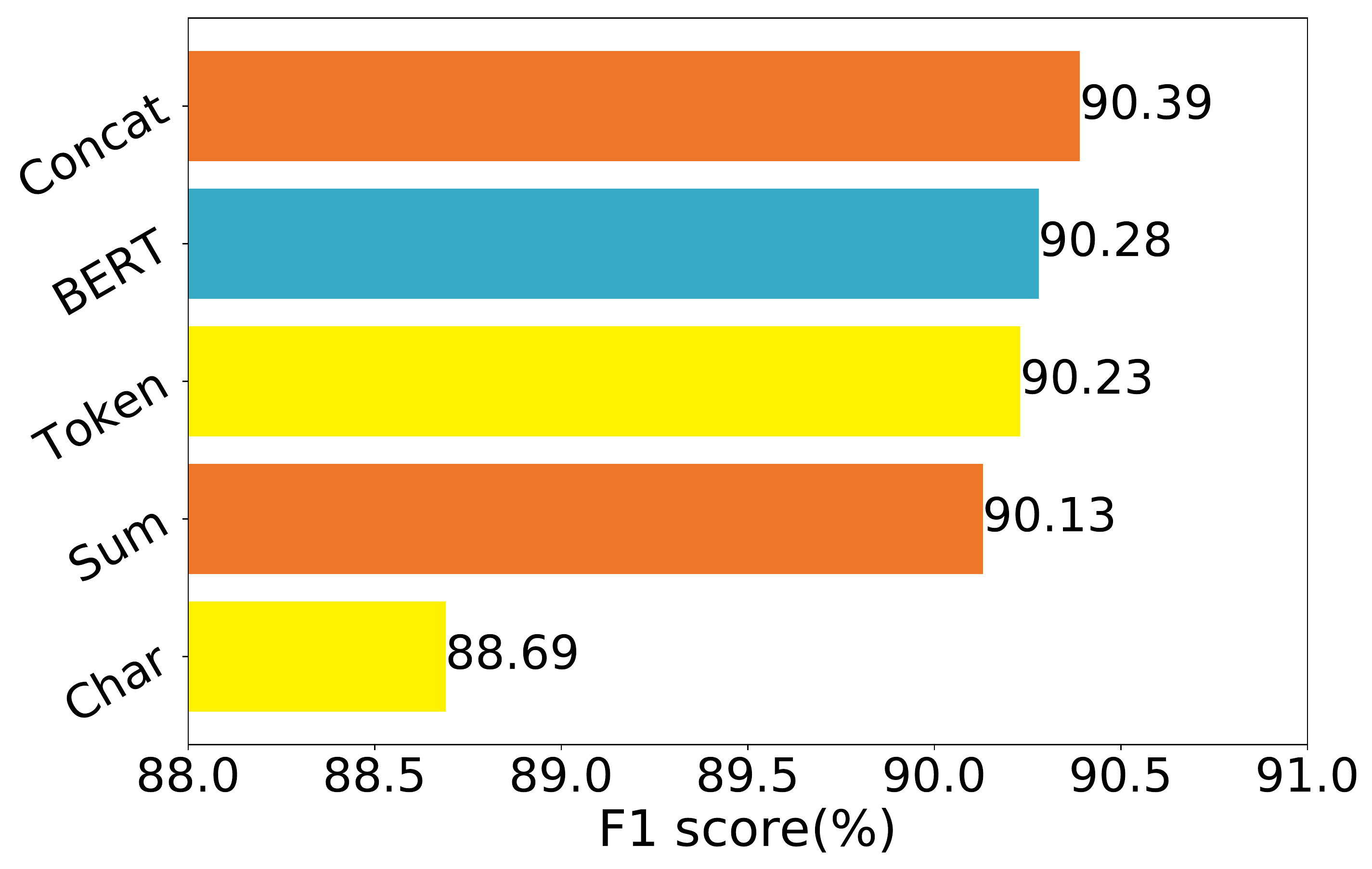} 
\caption{\label{result-embedding} Embedding comparison with LSTM on the CoNLL-2003 NER task. All the embeddings are generated by the hidden of the last layer. }
\end{minipage}
\end{figure}

We conduct the experiment with the original set and the set with character attacks.
For the words with multiple subwords, we use the hidden of the first subword as the word embedding, which is consistent with the fine-tuning setting.
For example, we calculate the sensitivity for each word in the sentence in the sample in Figure \ref {charbert-sen}, and the average of the results is $S$. 

To our surprise, the sensitivity results of the three models are: $S_{\texttt{BERT}} = 0.0612$, $S_{\texttt{AdvBERT}} = 0.0407$, $S_{\texttt{CharBERT}} = 0.0986$, but the robustness of the three models is: BERT \textless  AdvBERT \textless  CharBERT (Section \ref {sec: robustness}), which means there is no significant correlation between robustness and sensitivity. 
That is different from the previous work \newcite {pruthi-2019-combating}, which shows word recognition models with low sensitivity are more robust. After observing the results of many samples, we find that for the words without character noise, the sensitivity of BERT and CharBERT have no distinct difference. While for the words with noise such as `think-thnik,' `fair-far' in the example, the sensitivity of CharBERT is much higher than BERT. 
On the other hand, the sensitivity of AdvBERT is lower than BERT in most of the words.

That indicates CharBERT improves the robustness using a different way with adversarial training (AdvBERT).
It may be because we use the representations of noisy words to predict the original word in NLM, but AdvBERT treats all the words in the same way in the pre-training step, which leads CharBERT to construct the representations for the noisy words in a different way.
The result inspires us that, we can improve the robustness of model directly by better representation for the noise, which is different from improving the robustness by additional word recognition modules or adversarial training.

\subsection{Feature-based Comparison} 
The contextual word embeddings from pre-trained models are usually used as input features in task-specific models.
To explore whether the character information can enrich the word representation, we evaluate the contextual embedding generated by BERT and CharBERT.
Following  \newcite{devlin-2019-bert}, we use the same input representation as Section \ref{sec:results-ner} without fine-tuning any parameters of BERT or CharBERT.
Those contextual embeddings are used as embedding features to a randomly initialized two-layer 768-dimensional Bi-LSTM before the classification layer.
For CharBERT, we consider embeddings from three sources: token channel, character channel, sum, and concatenating of the two channels.

The results are reported in Figure \ref{result-embedding}.
We find that the embeddings from the token channel of CharBERT and BERT have similar performances, which denotes that the token channel retrains the information in BERT.
The embeddings from the character channel have worse performance, which may be due to the fewer data and training steps for this part of parameters.
When we concatenate the embeddings from the token and character channels, the model gets the best score.
That indicates the character information can enrich the word embeddings, even with a lot fewer training data and steps.

\section{Conclusion} 
In this paper, we address the important limitations of current PLMs: incomplete modeling and lack of robustness. 
To tackle these problems, we proposed a new pre-trained model CharBERT by injecting the character-level information into PLMs.
We construct the representations from characters by sequential GRU layers and use a dual-channel architecture for the subword and character.
Furthermore, we propose a new pre-training task NLM for unsupervised character representation learning.
The experimental results show that CharBERT can improve both the performance and robustness of pre-trained models.

In the future, we will extend CharBERT to other languages to learn cross-lingual representations from character information. We believe that CharBERT can bring even more improvements to morphologically rich languages like Arabic, where subwords cannot adequately capture the morphological information. On the other hand, we will extend CharBERT to defense other kinds of noise, e.g., word-level, sentence-level noise, to improve the robustness of PLMs comprehensively.

 \section*{Acknowledgement}\label{Acknowledgement}
 We would like to thank all anonymous reviewers for their hard work on reviewing and providing valuable comments on our paper. This work was supported by the National Natural Science Foundation of China (NSFC) via grant 61976072, 61632011, and 61772153.

\bibliographystyle{coling}
\bibliography{coling2020}

\end{document}